\documentclass[sigconf]{acmart}
\usepackage{tablefootnote}
\urldef{\challengeWebsiteURL}\url{https://explainable-deepfake-detection.github.io/}
\urldef{\challengeLeaderboardURL}\url{https://www.codabench.org/competitions/16461/#/results-tab}
\urldef{\challengePaperURL}\url{https://arxiv.org/abs/2607.21007}
\newcommand{\compactURL}[1]{{\renewcommand{\UrlFont}{\rmfamily\scriptsize}#1}}

\AtBeginDocument{%
  }

\setcopyright{acmlicensed}

\copyrightyear{2026}
\acmYear{2026}
\setcopyright{cc}
\setcctype{by-nc-nd}
\acmConference[MM '26]{Proceedings of the 34th ACM International Conference
on Multimedia}{November 10--14, 2026}{Rio de Janeiro, Brazil}
\acmBooktitle{Proceedings of the 34th ACM International Conference on
Multimedia (MM '26), November 10--14, 2026, Rio de Janeiro, Brazil}
\acmDOI{10.1145/3767308.3838609}
\acmISBN{979-8-4007-2213-4/2026/11}

\begin{document}

\title{MSU Team at the Explainable Deepfake Detection Challenge 2026: Grounded Artifact Evidence for Deepfake Detection}

\author{Artem Filippov}
\affiliation{%
  \institution{Lomonosov Moscow State University}
  \city{Moscow}
  \country{Russia}}
\email{artyom.filippov@graphics.cs.msu.ru}

\author{Aleksandr Gushchin}
\affiliation{%
  \institution{MSU Institute for Artificial Intelligence}
  \city{Moscow}
  \country{Russia}}
\email{alexander.gushchin@graphics.cs.msu.ru}

\author{Kirill Koltsov}
\affiliation{%
  \institution{Lomonosov Moscow State University}
  \city{Moscow}
  \country{Russia}}
\email{kirill.e.koltsov@mail.ru}

\author{Dmitriy Vatolin}
\affiliation{%
  \institution{MSU Institute for Artificial Intelligence}
  \city{Moscow}
  \country{Russia}}
\email{dmitriy@graphics.cs.msu.ru}

\author{Anastasia Antsiferova}
\affiliation{%
  \institution{MSU Institute for Artificial Intelligence}
  \city{Moscow}
  \country{Russia}}
\email{aantsiferova@graphics.cs.msu.ru}

\renewcommand{\shortauthors}{Artem Filippov, Aleksandr Gushchin, Kirill Koltsov, Dmitriy Vatolin, \& Anastasia Antsiferova}

\begin{abstract}
Recent advances in generative image models have made many manipulated images highly realistic, raising the need for detectors that are not only accurate but also able to provide visual evidence for their decisions. In this paper, we present our solution to the Explainable Deepfake Detection Challenge \cite{narang2026challenge} on the XPlainVerse dataset \cite{narang2026xplainverse}, where systems are required to predict whether an image is real or fake and generate both complex and simple explanations grounded in visible forensic cues.

Our method follows a modular detection-and-explanation design. For the real/fake decision, we build a multi-backbone detector that combines several DINOv3 models with Mesorch manipulation-localization features, bringing together pretrained visual representations, DCT-aware cues, and multi-scale forensic information. To inject explanation evidence into the detector, we use a Grounding-DINO-based pseudo-mask generation pipeline that converts local artifact descriptions from training explanations into weak patch-level supervision for an Artifact Evidence Map. We further introduce a local patch-level contrastive objective that separates artifact and authenticity evidence in the detector feature space without requiring paired images or pixel-level manipulation masks. For language output, we use class-conditional Qwen3-VL models to generate complex explanations for fake and real predictions, followed by a GRPO-optimized text simplification model.

The proposed methods were trained and evaluated on the challenge subset of XPlainVerse. On the full test split, our submission achieves 0.9349 detection accuracy, 0.5571 explanation score, and a 0.7456 final challenge score.
\end{abstract}



\maketitle

\section{Introduction}

Recent progress in generative image models has made many manipulated images increasingly realistic and harder to verify. A useful deepfake detector should therefore return not only a binary authenticity label but also evidence supporting the prediction; visible cues help audit model decisions and understand failures.

The Explainable Deepfake Detection Challenge \cite{narang2026challenge} directly targets this setting. It is built on XPlainVerse \cite{narang2026xplainverse}, a large-scale dataset of real and manipulated images with image-level labels, complex explanations for technical users, and simple explanations for general users. The official score combines real/fake classification with semantic similarity, simplicity, and LLM-based grounding metrics that check whether explanations refer to the correct entities and visual evidence. This encourages methods that connect the authenticity decision to the image cues described in the explanations.

Our solution separates real/fake detection from explanation generation. For the detector, we follow recent robust AI-generated image detection challenge results, where strong submissions often use ensembles of large pretrained visual backbones, especially DINOv3-based models \cite{Gushchin_2026_CVPR,simeoni2025dinov3}. We use several DINOv3 backbones and complement them with Mesorch features \cite{zhu2025mesoscopic}, which provide DCT-aware, multi-scale forensic representations.

To pass artifact information into the detector, we convert local artifact cues from fake-image explanations into weak patch-level pseudo-masks with a Grounding-DINO-based pipeline \cite{liu2024grounding,long2023opengroundingdino}. These masks encourage the detector to use visible, semantically grounded manipulation evidence when available.

We further adapt contrastive learning to the patch level, following the motivation of DRCT \cite{chen2024drct} and HiDA-Net \cite{mu2026no}: image-level contrastive objectives can improve fake-image separation, while local predictions encourage attention to fine forensic details. Our objective operates on patch embeddings and separates local fake and authentic evidence without paired images or pixel-level manipulation masks.

For explanation generation, we use class-conditional Qwen3-VL models \cite{bai2025qwen3} selected by the detector prediction, followed by a text-only simplification model optimized with GRPO \cite{shao2024deepseekmath}. This keeps classification in a specialized detector and uses VLMs for the challenge language outputs.

Our main contributions are \textbf{(I)} a pseudo-mask pipeline that extracts local forensic cues from complex explanations and grounds the corresponding image regions to obtain weak patch-level supervision, and \textbf{(II)} an explainable deepfake detection system combining DINOv3 and Mesorch features, patch-level artifact supervision, local contrastive learning, and class-conditional Qwen3-VL explanation generation.

\section{Related Work}

\subsection{Datasets}

Image-level synthetic-image benchmarks mainly target real/fake classification. GenImage \cite{zhu2023genimage} pairs ImageNet real images with GAN- and diffusion-generated counterparts. WildFake \cite{hong2025wildfake} contains over 3.5M real and fake images and organizes generators by model family, architecture, weights, and version, while Community Forensics \cite{park2025community} collects 2.7M images from 4,803 text-to-image models. These datasets are valuable for cross-generator generalization, but they mostly contain fully synthetic images and lack the explanation annotations required here.

MultiFakeVerse \cite{gupta2025multiverse} is closer to realistic misinformation scenarios, with 86,952 real and 758,041 person-centric manipulations involving people, objects, scenes, actions, and human--object interactions. Recent datasets also move toward explainable detection. DD-VQA \cite{zhang2024common} reformulates face forgery detection over FaceForensics++ \cite{rossler2019faceforensics} as visual question answering. FakeClue, introduced with FakeVLM \cite{wen2026spot}, contains more than 100K real and synthetic images with natural-language artifact explanations. LOKI \cite{ye2025loki} evaluates multimodal models across images, video, 3D, text, and audio using judgment, multiple-choice, abnormal-detail, and abnormal-explanation questions. Their formats, annotation protocols, and artifact distributions nevertheless differ from the challenge data.

\subsection{Methods}

AI-generated image detectors use reconstruction cues, as in AEROBLADE \cite{ricker2024aeroblade}; CLIP-based adaptation, as in FatFormer \cite{liu2024forgery}; or frequency information, as in SPAI \cite{karageorgiou2025any}. Recent challenge results show the practical strength of large pretrained backbones and ensembles. The NTIRE 2026 Robust AI-Generated Image Detection Challenge \cite{Gushchin_2026_CVPR} reports strong systems based on transformer ensembles, DINOv3 models \cite{simeoni2025dinov3}, robustness augmentation, and fusion, motivating our multi-backbone design.

DRCT \cite{chen2024drct} applies image-level contrastive learning to separate real, generated, and diffusion-reconstructed samples. HiDA-Net \cite{mu2026no} emphasizes spatial sensitivity to local edits, showing that local predictions can force attention to subtle forensic details. These ideas motivate our dense features and patch-level objective in addition to global classification.

Manipulation localization is especially relevant when only a small region is fake. TruFor \cite{guillaro2023trufor} combines RGB evidence with noise-sensitive representations, CAT-Net \cite{kwon2022learning} exploits RGB-DCT compression cues, and Mesorch \cite{zhu2025mesoscopic} combines DCT-aware inputs with multi-scale forensic features. We include Mesorch because its frequency-aware, localization-oriented representation complements DINOv3.

Vision-language models are increasingly used for explainable fake-image detection. FakeVLM \cite{wen2026spot} fine-tunes a multimodal model, including its visual encoder, for detection and artifact explanation. ForenDeX \cite{tan2026forendex} injects detector-guided forensic features into an MLLM, while FakeXplainer \cite{ji2026fakexplain} combines visual-encoder adaptation, supervised fine-tuning, and GRPO. These works show that VLMs are useful for explanation, but their visual components often need forensic adaptation; we therefore separate a specialized detector from class-conditional explanation generators.

\section{XPlainVerse Dataset and Challenge Subset}

XPlainVerse\cite{narang2026xplainverse} is the largest dataset for explainable deepfake detection, containing one million real and manipulated images paired with natural-language explanations. A substantial portion of it is based on the person-centric MultiFakeVerse dataset\cite{gupta2025multiverse}. The Explainable Deepfake Detection Challenge\cite{narang2026challenge} uses a curated 760,000-image subset: 270,000 real and 490,000 fake images. The public train, validation, and test splits contain 450,000, 110,000, and 200,000 images, respectively; labeled train--validation data comprise 180,000 real and 380,000 fake images, and the test split contains 90,000 real and 110,000 fake images.

\section{Methodology}

\begin{figure*}[t]
  \centering
  \includegraphics[width=\textwidth]{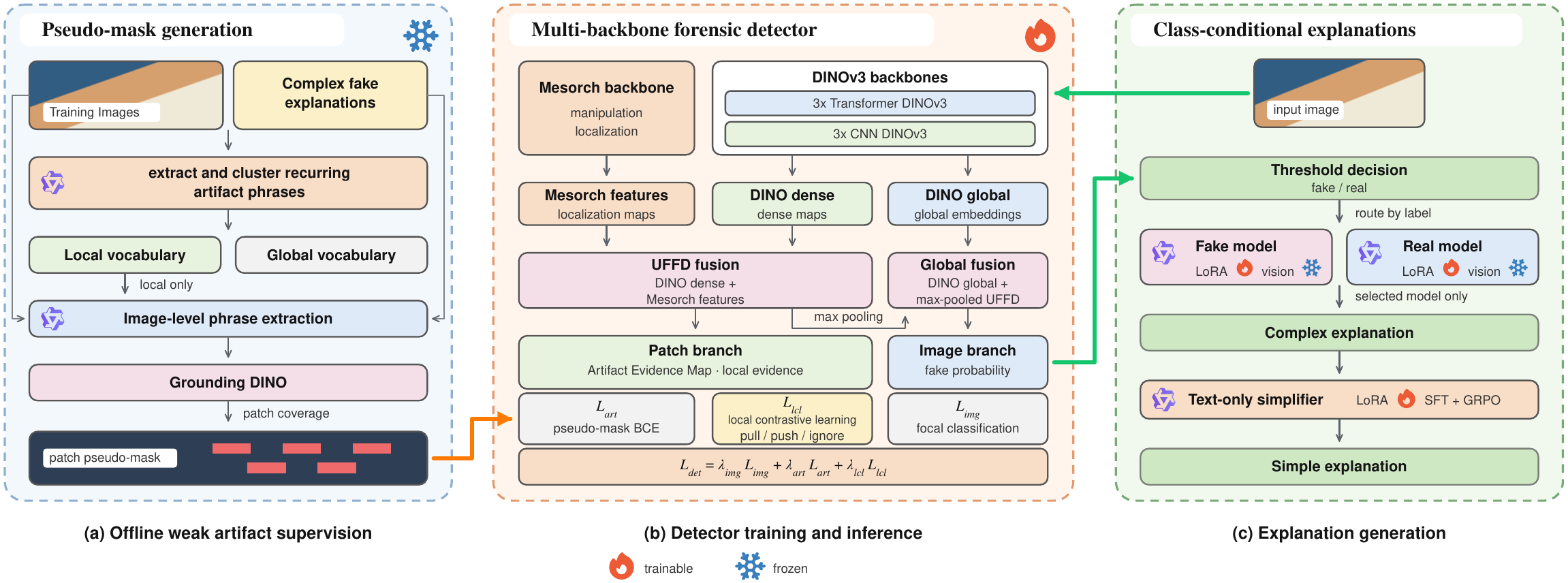}
  \caption{Overview of the proposed system, including offline pseudo-mask generation, multi-backbone forensic detection, and class-conditional explanation generation.}
  \Description{A three-part pipeline: explanation-grounded pseudo-mask generation, a multi-backbone detector with image and patch losses, and class-conditional complex and simple explanation generation.}
\end{figure*}

\subsection{Pseudo-mask Generation Pipeline}

We convert fake-image explanations into patch-level pseudo-masks used as weak artifact supervision. The masks are not pixel-accurate manipulation annotations; they indicate regions where textual evidence and open-vocabulary grounding agree.

Let $I_i$ have image-level label $y_i\in\{0,1\}$, where $y_i=1$ denotes fake. For fake images, the dataset provides a complex explanation $e_i$ describing the forensic evidence; we use it to derive a pseudo-mask $T_i\in[0,1]^{H_p\times W_p}$ on the detector patch grid. We do not extract artifact regions for real images and set $T_i\equiv0$. The pipeline runs offline for the training and validation splits and has vocabulary-construction and image-level grounding stages.

\subsubsection{Artifact Vocabulary Construction}

We first use Qwen3-VL-32B-Instruct \cite{bai2025qwen3} on a subset of image--explanation pairs to extract short artifact phrases from each complex explanation. The same model then merges near-duplicates and selects one representative per cluster. The resulting vocabulary is $\mathcal V=\{c_1,\ldots,c_K\}$, where each $c_k$ denotes a recurring artifact category observed in the dataset.

The normalized vocabulary is divided into disjoint local and global subsets, $\mathcal V=\mathcal V_{\mathrm{loc}}\cup\mathcal V_{\mathrm{glob}}$ and $\mathcal V_{\mathrm{loc}}\cap\mathcal V_{\mathrm{glob}}=\varnothing$. Local attributes can be associated with bounded regions, whereas global attributes describe image-wide properties. Only local attributes are grounded, avoiding noisy box supervision for global observations.

\subsubsection{Image-level Grounding Phrase Extraction}

For each fake image $I_i$, Qwen3-VL-32B-Instruct receives the image, its complex explanation $e_i$, and both attribute lists. It selects visually localizable artifacts mentioned or implied by the explanation and produces phrases for Grounding DINO \cite{liu2024grounding,long2023opengroundingdino}. Formally, it yields $\mathcal A_i=\{a_{i1},\ldots,a_{in_i}\}$ with $a_{ik}\in\mathcal V_{\mathrm{loc}}$ and one phrase $p_{ik}=g(a_{ik},I_i,e_i)$ per selected attribute. The phrases are concrete visual descriptions, such as an object part associated with the artifact, rather than abstract artifact names.

\subsubsection{Grounding and Patch Targets}

For each phrase $p_{ik}$, Grounding DINO returns candidate bounding boxes with confidence scores. Boxes below the confidence threshold are discarded, and all retained boxes are collected in $\mathcal B_i=\{b_{i1},\ldots,b_{im_i}\}$. We merge them into one artifact-evidence support because different artifact categories may overlap.

Let the detector grid have size $H_p\times W_p$, and let $R_{uv}$ denote the image region corresponding to patch $(u,v)$. The union $U_i=\bigcup_{b\in\mathcal B_i}b$ defines the grounded support, and the target is the covered fraction of each patch:
\[
t_{iuv}=\frac{|R_{uv}\cap U_i|}{|R_{uv}|},\qquad T_i[u,v]=t_{iuv},\qquad T_i\in[0,1]^{H_p\times W_p}.
\]
Thus, overlapping boxes contribute only once. The image-level label remains the primary supervision; $T_i$ only regularizes the patch-level artifact branch.

\subsection{Detector}

The detector predicts an image-level fake probability and a patch-level Artifact Evidence Map, and exposes patch embeddings for the local contrastive loss. For image $I_i$ with label $y_i$, it outputs an image-level logit $z_i$, an Artifact Evidence Map $A_i$, and patch embeddings from a shared forensic feature map.

\subsubsection{Multi-backbone Features}

We combine three transformer-based DINOv3 models \cite{simeoni2025dinov3}, three CNN-based DINOv3 models, and a Mesorch manipulation-localization backbone \cite{zhu2025mesoscopic}. For every DINOv3 backbone, we extract a global image embedding and the last dense feature map:
\[
f_i^{(k)}\in\mathbb R^{d_k},\qquad H_i^{(k)}\in\mathbb R^{H_p\times W_p\times c_k}.
\]
The DINO spatial maps are fused by per-backbone projection, concatenation, and an MLP over channels:
\[
H_i^{\mathrm{DINO}}=F_{\mathrm{map}}^{\mathrm{DINO}}\!\left(H_i^{(1)},\ldots,H_i^{(K)}\right),\qquad H_i^{\mathrm{DINO}}\in\mathbb R^{H_p\times W_p\times D}.
\]
The global embeddings are fused analogously:
\[
f_i^{\mathrm{DINO}}=F_{\mathrm{glob}}^{\mathrm{DINO}}\!\left(f_i^{(1)},\ldots,f_i^{(K)}\right),\qquad f_i^{\mathrm{DINO}}\in\mathbb R^D.
\]
We use Mesorch only as a feature extractor: its spatial representation is aligned to the DINO patch grid and projected to dimension $D$,
\[
H_i^{\mathrm{Mes}}=F_{\mathrm{Mes}}(I_i),\qquad H_i^{\mathrm{Mes}}\in\mathbb R^{H_p\times W_p\times D}.
\]

\subsubsection{Unified Forensic Feature Map}

The DINO and Mesorch spatial representations are fused into a shared patch-level representation called the \textbf{Unified Forensic Feature Map}. At each location $(u,v)$, we concatenate the corresponding vectors and apply an MLP:
\[
U_i[u,v]=F_{\mathrm{UFFM}}\!\left(H_i^{\mathrm{DINO}}[u,v],H_i^{\mathrm{Mes}}[u,v]\right),\qquad U_i\in\mathbb R^{H_p\times W_p\times D_U}.
\]

\subsubsection{Image-level Classification Branch}

The global branch predicts whether the image is fake. We summarize the Unified Forensic Feature Map with max pooling,
\[
u_i^{\mathrm{pool}}=\max_{u,v}U_i[u,v].
\]
Max pooling allows localized evidence to affect the image-level decision. The pooled descriptor is concatenated with the fused DINO global embedding, $g_i=[u_i^{\mathrm{pool}};f_i^{\mathrm{DINO}}]$, and passed through an MLP fusion block and binary classifier:
\[
z_i=h_{\mathrm{cls}}\!\left(F_{\mathrm{img}}(g_i)\right),\qquad p_i=\sigma(z_i).
\]
Here $p_i$ is the predicted probability that $I_i$ is fake. We train the branch with focal loss using $\alpha=0.5$ and $\gamma=2.0$.

\subsubsection{Artifact Evidence Map}

A shared patch head is applied to the Unified Forensic Feature Map:
\[
a_i[u,v]=h_{\mathrm{art}}\!\left(U_i[u,v]\right),\qquad q_i[u,v]=\sigma(a_i[u,v]).
\]
The logits $a_i$ form the Artifact Evidence Map $A_i$, and $q_i[u,v]$ estimates the probability that patch $(u,v)$ contains artifact evidence. The branch is supervised by $T_i$ with BCE; real-image targets are zero:
\[
\mathcal L_{\mathrm{art}}=\frac{1}{H_pW_p}\sum_{u,v}\operatorname{BCE}\!\left(q_i[u,v],T_i[u,v]\right).
\]

\subsubsection{Local Patch-level Contrastive Objective}

We use a local contrastive objective over patch embeddings from the Unified Forensic Feature Map. It organizes the patch space so that evidence consistent with fake images becomes more separable from authentic evidence, while uncertain regions are handled conservatively.

Batches contain fake--fake, real--real, and fake--real pairs. For each image pair $(I_a,I_b)$, patches are matched by cosine similarity. For $U_a[u,v]$, the matched patch in $I_b$ is
\[
(u^*,v^*)=\arg\max_{u',v'}\cos\!\left(U_a[u,v],U_b[u',v']\right),
\]
and matching is performed in both directions so that visually similar local regions can be compared even when they occur at different positions.

Each patch receives a status from the image-level label and current Artifact Evidence Map prediction $q_i[u,v]$. With uncertainty band $\delta$ around $0.5$,
\[
\begin{aligned}
|q_i[u,v]-0.5|\le\delta &\Rightarrow \mathrm{UF}\ (y_i=1),\ \mathrm{UR}\ (y_i=0),\\
q_i[u,v]>0.5+\delta &\Rightarrow \mathrm{CF}\ (y_i=1),\ \mathrm{WR}\ (y_i=0),\\
q_i[u,v]<0.5-\delta &\Rightarrow \mathrm{WF}\ (y_i=1),\ \mathrm{CR}\ (y_i=0).
\end{aligned}
\]
The suffix $F/R$ denotes whether the patch comes from a fake or real image, while $C/U/W$ denotes consistent, uncertain, or opposite-side local evidence. In other words, confident local predictions agreeing with the image label are treated as consistent, predictions near $0.5$ as uncertain, and confident contradictions as opposite-side evidence. Statuses are based on $q_i$ rather than $T_i$ because pseudo-masks may be incomplete.

For every matched patch pair, Tables~\ref{tab:fakefake}--\ref{tab:fakereal} select pull, push, or ignore. Tables~\ref{tab:fakefake} and \ref{tab:realreal} cover same-class image pairs; in Table~\ref{tab:fakereal}, rows correspond to fake-image patches and columns to real-image patches. Pull minimizes cosine distance,
\[
\ell_{\mathrm{pull}}(a,b)=1-\cos(a,b),
\]
while push separates embeddings with margin $m$,
\[
\ell_{\mathrm{push}}(a,b)=\max\!\left(0,m-[1-\cos(a,b)]\right).
\]
Ignored pairs do not contribute, and $\operatorname{sg}(\cdot)$ denotes stop-gradient.

\begin{table}[t]
  \caption{Local contrastive rules for fake--fake patch pairs.}
  \label{tab:fakefake}
  \begin{tabular}{lccc}
    \toprule
     & CF & UF & WF \\
    \midrule
    CF & pull(CF, CF) & pull(sg(CF), UF) & -- \\
    UF & pull(UF, sg(CF)) & pull(UF, UF) & -- \\
    WF & -- & -- & -- \\
    \bottomrule
  \end{tabular}
\end{table}

\begin{table}[t]
  \caption{Local contrastive rules for real--real patch pairs.}
  \label{tab:realreal}
  \begin{tabular}{lccc}
    \toprule
     & CR & UR & WR \\
    \midrule
    CR & pull(CR, CR) & pull(sg(CR), UR) & pull(sg(CR), WR) \\
    UR & pull(UR, sg(CR)) & pull(UR, UR) & pull(sg(UR), WR) \\
    WR & pull(WR, sg(CR)) & pull(WR, sg(UR)) & -- \\
    \bottomrule
  \end{tabular}
\end{table}

\begin{table}[t]
  \caption{Local contrastive rules for fake--real patch pairs.}
  \label{tab:fakereal}
  \begin{tabular}{lccc}
    \toprule
     & CR & UR & WR \\
    \midrule
    CF & push(CF, CR) & push(sg(CF), UR) & push(sg(CF), WR) \\
    UF & push(UF, sg(CR)) & push(UF, UR) & pull(sg(UF), WR) \\
    WF & -- & pull(sg(WF), UR) & pull(sg(WF), WR) \\
    \bottomrule
  \end{tabular}
\end{table}

The rules treat confident fake artifacts and confident real evidence as reliable anchors. Ambiguous or opposite-side patches are either updated through stop-gradient anchors or ignored, preventing uncertain matches from moving both embeddings simultaneously. The local loss over all retained matches is
\[
\mathcal L_{\mathrm{lcl}}=\sum_{(a,b)}\ell_{ab},
\]
where $\ell_{ab}$ is the selected pull or push term.

\subsubsection{Training Objective}

The final detector objective combines image classification, artifact-map supervision, and local contrastive learning:
\[
\mathcal L_{\mathrm{det}}=\lambda_{\mathrm{img}}\mathcal L_{\mathrm{img}}+\lambda_{\mathrm{art}}\mathcal L_{\mathrm{art}}+\lambda_{\mathrm{lcl}}\mathcal L_{\mathrm{lcl}}.
\]
The terms supervise image classification, weak artifact localization, and patch-space organization.

\begin{figure*}[t]
  \centering
  \includegraphics[width=\textwidth]{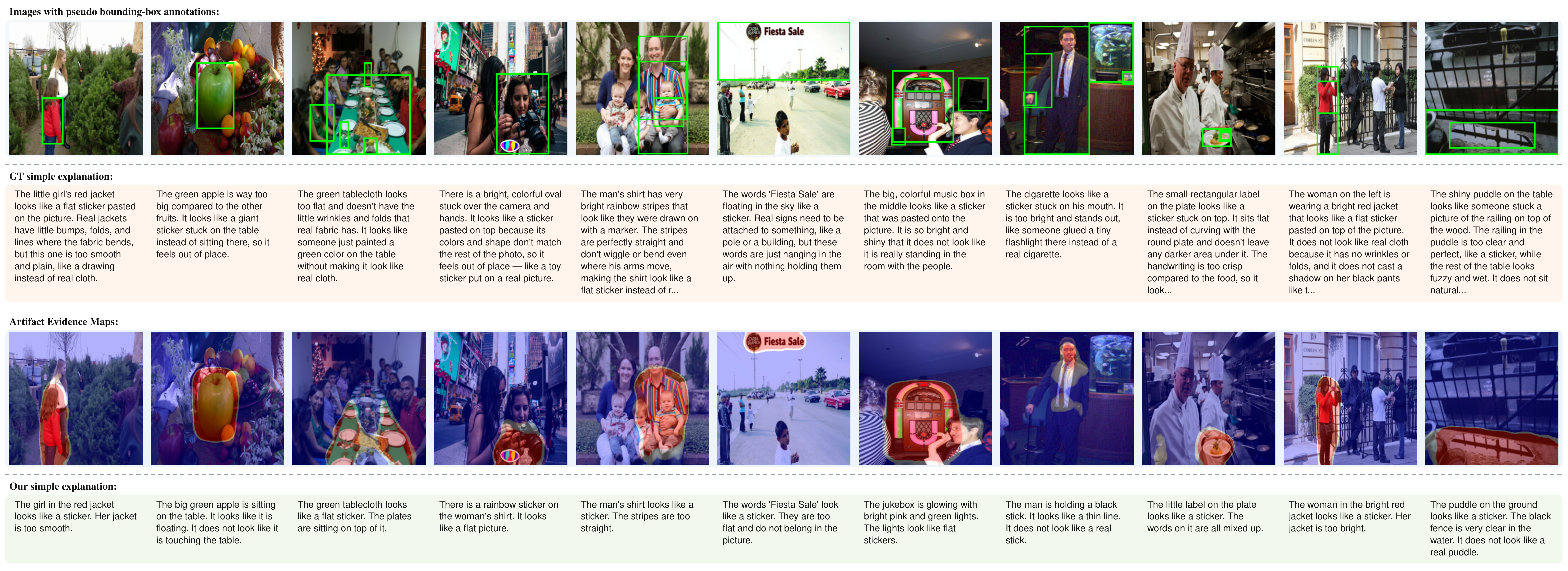}
  \caption{Internal-validation examples. From top to bottom: pseudo boxes, ground-truth simple explanations, Artifact Evidence Maps, and generated simple explanations.}
  \Description{Eleven examples showing grounded pseudo boxes, reference simple explanations, detector evidence heatmaps, and generated simple explanations.}
  \label{fig:dataset-examples}
\end{figure*}

\subsection{Class-Conditional Vision-Language Explanation Generation}

Natural-language explanations are generated by separate Qwen3-VL-8B-Instruct models \cite{bai2025qwen3} selected by the detector prediction. The complex explanation is the evidence-bearing output, while the simple explanation is derived afterward for general users. The module uses $G_{\mathrm{fake}}$ and $G_{\mathrm{real}}$ for complex explanations and a text-only simplifier $S$.

\subsubsection{Class-conditional Complex Explanation Models}

The complex explanation is the primary evidence-bearing text output. We train separate generators for fake and real images to reduce class leakage, especially hallucinated artifacts for real images:
\[
G_{\mathrm{fake}}:I_i\mapsto\hat e_{i,\mathrm{fake}}^c,\qquad G_{\mathrm{real}}:I_i\mapsto\hat e_{i,\mathrm{real}}^c.
\]
At inference, only the generator selected by the detector prediction is used.

\subsubsection{Supervised Fine-tuning}

Both generators are LoRA-adapted from Qwen3-VL-8B-Instruct. LoRA is applied to all linear layers, while the vision tower is frozen to reduce memory cost and preserve the pretrained visual representation. We train on the public labeled data and reserve an internal subset for model selection and qualitative checking. Because real images are less frequent, they are oversampled when training the real-explanation branch.

\subsubsection{Complex-to-simple Explanation Model}

The challenge also requires an explanation for general users. We train a text-only simplifier, $\hat e_i^s=S(\hat e_i^c)$, which makes inference and GRPO rollouts cheaper. Real-image complex and simple explanations are often nearly identical, so fake-image pairs provide most of the simplification signal. After supervised fine-tuning, $S$ is optimized with GRPO \cite{shao2024deepseekmath} using the official simple-explanation score.

\subsubsection{Inference Pipeline}

At inference, the detector first predicts $\hat y_i=\mathbf 1[p_i\ge\tau]$, where $\tau$ is selected on validation data. This label routes the image to exactly one complex-explanation model:
\[
\hat e_i^c=\begin{cases}
G_{\mathrm{fake}}(I_i), & \hat y_i=1,\\
G_{\mathrm{real}}(I_i), & \hat y_i=0,
\end{cases}
\qquad \hat e_i^s=S(\hat e_i^c).
\]
The selected generator produces the complex explanation, which is then passed to the simplifier. The final output is $(\hat y_i,\hat e_i^c,\hat e_i^s)$: the detector supplies the label, the selected VLM the complex explanation, and $S$ the simple explanation.

\section{Experiments}

\subsection{Detector Ablation}

All detector ablations were trained on the public training split and evaluated on the public validation split. Each model was trained for 10 epochs with per-device batch size 4, gradient accumulation over 16 steps, learning rate $5\times10^{-4}$, cosine decay, warmup ratio 0.03, and maximum gradient norm 1.0. A fixed augmentation pipeline used random 90-degree rotations, horizontal and vertical flips, random resized crop, bicubic resizing, JPEG compression, additive white noise, impulse noise, brightness changes, darkening, and color quantization. Only the three loss weights differed; we report validation AUC for image-level prediction.

\begin{table}[t]
  \caption{Detector ablation on the public validation split.}
  \label{tab:detector-ablation}
  \centering
  \small
  \setlength{\tabcolsep}{3.2pt}
  \begin{tabular}{lcccc}
    \toprule
    Variant & $\lambda_{\mathrm{img}}$ & $\lambda_{\mathrm{art}}$ & $\lambda_{\mathrm{lcl}}$ & Val. AUC \\
    \midrule
    Image-level detector only & 15 & 0 & 0 & 0.8975 \\
    + pseudo-mask artifact supervision & 15 & 1 & 0 & 0.9139 \\
    + LCL & 15 & 1 & 1 & \textbf{0.9417} \\
    \bottomrule
  \end{tabular}
\end{table}

\begin{table*}[!t]
  \caption{Explanation generation evaluation on the internal held-out validation subset. $B_c$ and $B_s$ denote complex and simple BERTScore-F1; $F_{\mathrm{ent}}$ and $F_{\mathrm{evid}}$ denote Entity and Evidence F1; $L_s$, $S_s$, and $M_{\mathrm{exp}}$ denote normalized simple SLE, Simple score, and Explanation score.}
  \label{tab:explanation-eval}
  \centering
  \footnotesize
  \setlength{\tabcolsep}{4pt}
  \renewcommand{\arraystretch}{0.94}
  \begin{tabular*}{\textwidth}{@{\extracolsep{\fill}}lccccccc@{}}
    \toprule
    Method & $B_c$ & $F_{\mathrm{ent}}$ & $F_{\mathrm{evid}}$ & $B_s$ & $L_s$ & $S_s$ & $M_{\mathrm{exp}}$ \\
    \midrule
    Zero-shot Qwen3-VL-253B-A22B-Instruct & 0.6023 & 0.5310 & 0.3934 & 0.4495 & 0.3418 & 0.4172 & 0.4812 \\
    Organizer baseline\tablefootnote{The organizer baseline model can be found on the official challenge website:\newline\compactURL{\challengeWebsiteURL}} & 0.7109 & 0.5536 & 0.4620 & 0.4578 & 0.4264 & 0.4484 & 0.5365 \\
    Ours & \textbf{0.7253} & \textbf{0.6235} & \textbf{0.5447} & \textbf{0.4991} & \textbf{0.9703} & \textbf{0.6405} & \textbf{0.6236} \\
    \bottomrule
  \end{tabular*}
\end{table*}

Adding the Artifact Evidence Map improves AUC over the image-level detector, from 0.8975 to 0.9139. This suggests that weak supervision derived from explanation-grounded boxes helps the detector learn localized artifact cues consistent with the textual evidence. The local contrastive loss provides the largest additional gain, raising AUC to 0.9417. This is consistent with its purpose: organizing patch embeddings around local artifact and authenticity evidence can improve the final image-level decision.

For the final submissions, the detector was trained on all available public labeled data rather than only the original training split. Both the pseudo-mask and local contrastive losses were retained.

\subsection{Explanation Generation Evaluation}

We evaluate explanation generation on an internal held-out subset of 10K examples, consisting of 5K real and 5K fake images, using the official evaluation script \cite{narang2026challenge}. The metrics include BERTScore-F1 \cite{zhang2019bertscore} for complex and simple explanations, normalized SLE \cite{cripwell2023simplicity}, entity and evidence F1, the simple score, and the final explanation score.

To isolate explanation quality from detector errors, this evaluation uses only generations with the correct final answer. Zero-shot Qwen3-VL uses class-specific prompts, the organizer baseline uses its provided answer-prefill format, and our method uses separate fake and real Qwen3-VL-8B complex-explanation models followed by the GRPO-optimized simplifier.

All models use rank-16 LoRA with alpha 32 and all-linear targets. The complex generators and supervised simplifier are trained for 4 epochs with effective batch size 128 and learning rate $2\times10^{-4}$. GRPO runs for 1 epoch with effective batch size 32, learning rate $10^{-4}$, 32 sampled generations per prompt, and KL coefficient $\beta=0.01$.

Our explanation pipeline improves all internal metrics. The largest gain is in normalized SLE, from 0.4264 for the organizer baseline to 0.9703, while simple BERTScore also improves. This suggests that direct optimization of the official objective produces shorter explanations while preserving semantic alignment. Class-conditional generators raise entity F1 from 0.5536 to 0.6235 and evidence F1 from 0.4620 to 0.5447, consistent with improved grounding specialization.

\subsection{Final Submission Results}

We report results on the full final test split, including the detector output and both generated explanations used in the official evaluation \cite{narang2026challenge}.

\begin{table}[!t]
  \caption{Final submission results on the full final test split.}
  \label{tab:final-results}
  \centering
  \footnotesize
  \renewcommand{\arraystretch}{0.90}
  \begin{tabular}{lc}
    \toprule
    Metric & Score \\
    \midrule
    Detection accuracy & 0.9349 \\
    Detection fake F1 & 0.9418 \\
    Detection macro F1 & 0.9340 \\
    Detection real F1 & 0.9261 \\
    Complex BERTScore-F1 & 0.7004 \\
    Simple BERTScore-F1 & 0.6509 \\
    Normalized SLE & 0.9726 \\
    Entity score & 0.4987 \\
    Evidence score & 0.3932 \\
    Explanation score & 0.5571 \\
    Final challenge score\tablefootnote{The official challenge leaderboard is available at:\newline\compactURL{\challengeLeaderboardURL}}$^,$\tablefootnote{Extended metrics for the top-5 teams are reported in the challenge paper~\cite{narang2026challenge}.} & \textbf{0.7456} \\
    \bottomrule
  \end{tabular}
\end{table}

The submission reaches 0.9349 detection accuracy and 0.9340 macro F1, with fake and real F1 scores of 0.9418 and 0.9261. Its explanation score is 0.5571, and the combined final challenge score is 0.7456.

\section{Discussion}

The proposed system achieved strong final results in the challenge setting, but it is computationally heavy. It combines several DINOv3 backbones, Mesorch features, and multiple Qwen3-VL models. This complexity is acceptable for a challenge submission, but it makes the method expensive and less attractive for practical deployment.

A key limitation is that the LLM/VLM component is used mostly as an attachment to the detector. The detector first makes the real/fake decision, and the class-conditional Qwen3-VL model then generates an explanation for the predicted class. This improves the challenge-format output, but it does not fully use the reasoning potential of multimodal language models: the VLM does not participate in the authenticity decision, compare alternative hypotheses, or verify detector evidence.

Although GRPO improves the official metrics for simple explanations, especially normalized SLE, this does not necessarily mean that they become substantially more useful or accessible to ordinary users.

\section{Conclusion}

We presented a modular solution for explainable deepfake detection trained and evaluated on the XPlainVerse challenge subset \cite{narang2026xplainverse}. The system combines a multi-backbone forensic detector, explanation-grounded pseudo-mask supervision, local patch-level contrastive learning, class-conditional Qwen3-VL explanation generation, and GRPO-based simplification. On the final test split, it achieved 0.9349 detection accuracy, a 0.5571 explanation score, and a 0.7456 final challenge score.

In validation, ablations indicate that local supervision derived from explanations improves detector generalization, with a further gain from local contrastive learning. GRPO-based simplification also improves the official simple-explanation score and contributes to the final result.

\begin{acks}
The research was carried out using the MSU-270 supercomputer of Lomonosov Moscow State University. 
\end{acks}


\clearpage
\begin{thebibliography}{99}

\bibitem{narang2026xplainverse}
Abhijeet Narang, Kartik Kuckreja, Shreya Ghosh, Muhammad Haris Khan, Jianfei Cai, and Abhinav Dhall. 2026.
\newblock XPlainVerse: A Million-Scale Benchmark for Explainable Deepfake Detection.
\newblock \emph{arXiv preprint arXiv:2607.03562}.
\newblock \url{https://arxiv.org/abs/2607.03562}

\bibitem{narang2026challenge}
Abhijeet Narang, Kartik Kuckreja, Shreya Ghosh, Muhammad Haris Khan, Usman Tariq, Jianfei Cai, and Abhinav Dhall. 2026.
\newblock Explainable Deepfake Detection Challenge.
\newblock \emph{arXiv preprint arXiv:2607.21007}.
\newblock \url{https://arxiv.org/abs/2607.21007}

\bibitem{Gushchin_2026_CVPR}
Aleksandr Gushchin, Khaled Abud, Ekaterina Shumitskaya, Artem Filippov, Georgii Bychkov, Sergey Lavrushkin, Mikhail Erofeev, Anastasia Antsiferova, Changsheng Chen, Shunquan Tan, Radu Timofte, Dmitriy Vatolin, Chuanbiao Song, Zijian Yu, Hao Tan, Jun Lan, Zhiqiang Yang, Yongwei Tang, Zhiqiang Wu, Jia Wen Seow, Hong Vin Koay, Haodong Ren, Feng Xu, Shuai Chen, Ruiyang Xia, Qi Zhang, Yaowen Xu, Zhaofan Zou, Hao Sun, Dagong Lu, Mufeng Yao, Xinlei Xu, Fei Wu, Fengjun Guo, Cong Luo, Hardik Sharma, Aashish Negi, Prateek Shaily, Jayant Kumar, Sachin Chaudhary, Akshay Dudhane, Praful Hambarde, Amit Shukla, Zhilin Tu, Fengpeng Li, Jiamin Zhang, Jianwei Fei, Kemou Li, Haiwei Wu, Bilel Benjdira, Anas M. Ali, Wadii Boulila, Chenfan Qu, and Junchi Li. 2026.
\newblock NTIRE 2026 Challenge on Robust AI-Generated Image Detection in the Wild.
\newblock In \emph{Proceedings of the IEEE/CVF Conference on Computer Vision and Pattern Recognition (CVPR) Workshops}, 1895--1913.

\bibitem{simeoni2025dinov3}
Oriane Sim{\'e}oni, Huy V. Vo, Maximilian Seitzer, Federico Baldassarre, Maxime Oquab, Cijo Jose, Vasil Khalidov, Marc Szafraniec, Seungeun Yi, Micha{\"e}l Ramamonjisoa, and others. 2025.
\newblock DINOv3.
\newblock \emph{arXiv preprint arXiv:2508.10104}.

\bibitem{zhu2025mesoscopic}
Xuekang Zhu, Xiaochen Ma, Lei Su, Zhuohang Jiang, Bo Du, Xiwen Wang, Zeyu Lei, Wentao Feng, Chi-Man Pun, and Ji-Zhe Zhou. 2025.
\newblock Mesoscopic insights: orchestrating multi-scale \& hybrid architecture for image manipulation localization.
\newblock In \emph{Proceedings of the AAAI Conference on Artificial Intelligence}, Vol. 39, No. 10, 11022--11030.

\bibitem{liu2024grounding}
Shilong Liu, Zhaoyang Zeng, Tianhe Ren, Feng Li, Hao Zhang, Jie Yang, Qing Jiang, Chunyuan Li, Jianwei Yang, Hang Su, and others. 2024.
\newblock Grounding DINO: Marrying DINO with grounded pre-training for open-set object detection.
\newblock In \emph{European Conference on Computer Vision}, Springer, 38--55.

\bibitem{long2023opengroundingdino}
Zuwei Long and Wei Li. 2023.
\newblock Open Grounding Dino: The third party implementation of the paper Grounding DINO.
\newblock \url{https://github.com/longzw1997/Open-GroundingDino}.

\bibitem{chen2024drct}
Baoying Chen, Jishen Zeng, Jianquan Yang, and Rui Yang. 2024.
\newblock DRCT: Diffusion reconstruction contrastive training towards universal detection of diffusion generated images.
\newblock In \emph{Forty-first International Conference on Machine Learning}.

\bibitem{mu2026no}
Lianrui Mu, Haoji Hu, Zou Xingze, Jianhong Bai, and Jiaqi Hu. 2026.
\newblock No Pixel Left Behind: A Detail-Preserving Architecture for Robust High-Resolution AI-Generated Image Detection.
\newblock In \emph{The Fourteenth International Conference on Learning Representations}.
\newblock \url{https://openreview.net/forum?id=9QQ3Kc2hj6}

\bibitem{bai2025qwen3}
Shuai Bai, Yuxuan Cai, Ruizhe Chen, Keqin Chen, Xionghui Chen, Zesen Cheng, Lianghao Deng, Wei Ding, Chang Gao, Chunjiang Ge, and others. 2025.
\newblock Qwen3-VL technical report.
\newblock \emph{arXiv preprint arXiv:2511.21631}.

\bibitem{shao2024deepseekmath}
Zhihong Shao, Peiyi Wang, Qihao Zhu, Runxin Xu, Junxiao Song, Xiao Bi, Haowei Zhang, Mingchuan Zhang, YK Li, Yang Wu, and others. 2024.
\newblock DeepSeekMath: Pushing the limits of mathematical reasoning in open language models.
\newblock \emph{arXiv preprint arXiv:2402.03300}.

\bibitem{zhu2023genimage}
Mingjian Zhu, Hanting Chen, Qiangyu Yan, Xudong Huang, Guanyu Lin, Wei Li, Zhijun Tu, Hailin Hu, Jie Hu, and Yunhe Wang. 2023.
\newblock GenImage: A million-scale benchmark for detecting AI-generated image.
\newblock \emph{Advances in Neural Information Processing Systems} 36 (2023), 77771--77782.

\bibitem{hong2025wildfake}
Yan Hong, Jianming Feng, Haoxing Chen, Jun Lan, Huijia Zhu, Weiqiang Wang, and Jianfu Zhang. 2025.
\newblock WildFake: A large-scale and hierarchical dataset for AI-generated images detection.
\newblock In \emph{Proceedings of the AAAI Conference on Artificial Intelligence}, Vol. 39, No. 4, 3500--3508.
\\ \eject
\bibitem{park2025community}
Jeongsoo Park and Andrew Owens. 2025.
\newblock Community forensics: Using thousands of generators to train fake image detectors.
\newblock In \emph{Proceedings of the Computer Vision and Pattern Recognition Conference}, 8245--8257.

\bibitem{gupta2025multiverse}
Parul Gupta, Shreya Ghosh, Tom Gedeon, Thanh-Toan Do, and Abhinav Dhall. 2025.
\newblock Multiverse Through Deepfakes: The MultiFakeVerse Dataset of Person-Centric Visual and Conceptual Manipulations.
\newblock In \emph{Proceedings of the 33rd ACM International Conference on Multimedia (MM '25)}, 13258--13265.
\newblock \url{https://doi.org/10.1145/3746027.3758283}

\bibitem{zhang2024common}
Yue Zhang, Ben Colman, Xiao Guo, Ali Shahriyari, and Gaurav Bharaj. 2024.
\newblock Common sense reasoning for deepfake detection.
\newblock In \emph{European Conference on Computer Vision}, Springer, 399--415.

\bibitem{rossler2019faceforensics}
Andreas R{\"o}ssler, Davide Cozzolino, Luisa Verdoliva, Christian Riess, Justus Thies, and Matthias Nie{\ss}ner. 2019.
\newblock FaceForensics++: Learning to detect manipulated facial images.
\newblock In \emph{Proceedings of the IEEE/CVF International Conference on Computer Vision}, 1--11.

\bibitem{wen2026spot}
Siwei Wen, Peilin Feng, Hengrui Kang, Zichen Wen, Yize Chen, Jiang Wu, Conghui He, Weijia Li, and others. 2026.
\newblock Spot the fake: Large multimodal model-based synthetic image detection with artifact explanation.
\newblock \emph{Advances in Neural Information Processing Systems} 38 (2026), 58972--59005.

\bibitem{ye2025loki}
Junyan Ye, Baichuan Zhou, Zilong Huang, Junan Zhang, Tianyi Bai, Hengrui Kang, Jun He, Honglin Lin, Zihao Wang, Tong Wu, and others. 2025.
\newblock LOKI: A comprehensive synthetic data detection benchmark using large multimodal models.
\newblock In \emph{International Conference on Learning Representations}, Vol. 2025, 70440--70522.

\bibitem{ricker2024aeroblade}
Jonas Ricker, Denis Lukovnikov, and Asja Fischer. 2024.
\newblock AEROBLADE: Training-free detection of latent diffusion images using autoencoder reconstruction error.
\newblock In \emph{Proceedings of the IEEE/CVF Conference on Computer Vision and Pattern Recognition}, 9130--9140.

\bibitem{liu2024forgery}
Huan Liu, Zichang Tan, Chuangchuang Tan, Yunchao Wei, Jingdong Wang, and Yao Zhao. 2024.
\newblock Forgery-aware adaptive transformer for generalizable synthetic image detection.
\newblock In \emph{Proceedings of the IEEE/CVF Conference on Computer Vision and Pattern Recognition}, 10770--10780.

\bibitem{karageorgiou2025any}
Dimitrios Karageorgiou, Symeon Papadopoulos, Ioannis Kompatsiaris, and Efstratios Gavves. 2025.
\newblock Any-resolution AI-generated image detection by spectral learning.
\newblock In \emph{Proceedings of the Computer Vision and Pattern Recognition Conference}, 18706--18717.

\bibitem{guillaro2023trufor}
Fabrizio Guillaro, Davide Cozzolino, Avneesh Sud, Nicholas Dufour, and Luisa Verdoliva. 2023.
\newblock TruFor: Leveraging all-round clues for trustworthy image forgery detection and localization.
\newblock In \emph{Proceedings of the IEEE/CVF Conference on Computer Vision and Pattern Recognition}, 20606--20615.

\bibitem{kwon2022learning}
Myung-Joon Kwon, Seung-Hun Nam, In-Jae Yu, Heung-Kyu Lee, and Changick Kim. 2022.
\newblock Learning JPEG compression artifacts for image manipulation detection and localization.
\newblock \emph{International Journal of Computer Vision} 130, 8 (2022), 1875--1895.

\bibitem{tan2026forendex}
Chuangchuang Tan, Jinglu Wang, Xiang Ming, Renshuai Tao, Yunchao Wei, Yao Zhao, and Yan Lu. 2026.
\newblock ForenDeX: Unlocking Forensic Insights for Explainable AI-Generated Image Detection.
\newblock In \emph{Proceedings of the IEEE/CVF Conference on Computer Vision and Pattern Recognition}, 6592--6601.

\bibitem{ji2026fakexplain}
Yikun Ji, Yan Hong, Qi Fan, Huijia Zhu, Weiqiang Wang, Liqing Zhang, Jianfu Zhang, and others. 2026.
\newblock FakeXplain: AI-generated image detection via human-aligned grounded reasoning.
\newblock In \emph{The Fourteenth International Conference on Learning Representations}.

\bibitem{zhang2019bertscore}
Tianyi Zhang, Varsha Kishore, Felix Wu, Kilian Q. Weinberger, and Yoav Artzi. 2019.
\newblock BERTScore: Evaluating text generation with BERT.
\newblock \emph{arXiv preprint arXiv:1904.09675}.

\bibitem{cripwell2023simplicity}
Liam Cripwell, Jo{\"e}l Legrand, and Claire Gardent. 2023.
\newblock Simplicity level estimate (SLE): A learned reference-less metric for sentence simplification.
\newblock In \emph{Proceedings of the 2023 Conference on Empirical Methods in Natural Language Processing}, 12053--12059.

\end{thebibliography}
\end{document}